\documentclass[preprint,12pt]{elsarticle}

\usepackage{amssymb}
\usepackage{amsmath}

\usepackage{booktabs}
\usepackage[accsupp]{axessibility}
\usepackage{xspace}
\newcommand{\ours}{RecOE\xspace}
\newcommand{\ood}{OoD\xspace}
\newcommand{\ind}{InD\xspace}
\usepackage{adjustbox}
\usepackage{multirow}
\usepackage{makecell}
\usepackage{tabularx}
\newcolumntype{C}{>{\centering\arraybackslash}X}
\newcolumntype{L}{>{\raggedright\arraybackslash}X}
\usepackage{array}
\usepackage{bm}  
\usepackage{threeparttable}
\usepackage{url}
\usepackage[table]{xcolor}
\usepackage{caption}
\definecolor{resultgray}{gray}{0.94}
\newcommand{\resulttable}{%
  \small\def\baselinestretch{1.05}\selectfont
  \setlength{\tabcolsep}{5pt}%
  \renewcommand{\arraystretch}{1.16}}
\newcommand{\resultrow}{\rowcolor{resultgray}}
\newcommand{\stat}[2]{#1\,{\scriptstyle\pm\,#2}}

\usepackage{algorithm}
\usepackage{algpseudocode}
\usepackage{fix-cm}

\begin{document}

\begin{frontmatter}



\title{Subspace Learning with Outlier Exposure for Out-of-Distribution Detection}




\author[aff1]{Hongxu Shen\fnref{equal}}
\ead{shenhongxu@bit.edu.cn}

\author[aff2]{Kun Fang\fnref{equal}}
\ead{kun.fang@polyu.edu.hk}

\author[aff3]{Jiahui Zheng}
\ead{zhengjh14@tsinghua.org.cn}

\author[aff1]{Jian Sun}
\ead{sunjian@bit.edu.cn}

\author[aff1]{Qinghua Tao\corref{cor1}}
\ead{qinghua.tao@bit.edu.cn}

\fntext[equal]{These authors contributed equally to this work.}

\cortext[cor1]{Corresponding author.}

\affiliation[aff1]{
    organization={School of Automation, Beijing Institute of Technology},
    city={Beijing},
    country={China}
}

\affiliation[aff2]{
    organization={The Hong Kong Polytechnic University},
    city={Hong Kong},
    country={China}
}

\affiliation[aff3]{
    organization={Tsinghua University},
    city={Beijing},
    country={China}
}

\begin{abstract}
Out-of-Distribution (\ood) detection is pivotal for reliable deployments of Deep Neural Networks (DNNs), aiming to distinguish anomalous samples from the In-Distribution (\ind) data in training.  
By introducing outliers as surrogate \ood data to finetune DNNs, Outlier Exposure (OE) has greatly advanced \ood detection in recent years. 
Existing OE-based research remains predominantly logit-centric, either by generating {diverse} outliers or by regularizing them with objectives operating at the prediction-output level.
Nonetheless, these approaches leave much of the rich information in feature spaces untapped.
In this work, we propose an OE-based detection framework, named \ours, that learns appropriate feature subspaces from \ind data and auxiliary outliers, such that the \ind-\ood discrepancy can be effectively characterized through the learned feature subspaces.
Specifically, the feature subspaces are optimized to capture the dominant variations of InD and outliers by minimizing reconstruction errors therein.
Besides, a sample-level objective discourages reconstruction in the opposing subspace, while an orthogonality regularizer reduces subspace overlap, further promoting  discriminative subspaces.
At inference, the difference in reconstruction errors between the learned subspaces serves as the \ood detection score. 
Extensive experiments show the superiority of \ours on multiple benchmarks, in which significant performance gains are achieved over recent strong baselines with only slight extra overhead from two small-size matrices of subspace bases.
The source code is available at
\url{https://github.com/shenhongxu123/RecOE}.
\end{abstract}




\end{frontmatter}



\section{Introduction}
\label{sec:intro}

While deep neural networks (DNNs) have achieved vast success in real-world applications \cite{goodfellow2016deep,gu2018recent}, they are known to yield overconfident predictions on samples that deviate from the in-distribution (\ind)  training data \cite{nguyen2015deep}, commonly known as Out-of-Distribution (\ood) samples.
Such behavior poses substantial risks in safety-critical scenarios, including autonomous driving, medical diagnosis, and financial systems \cite{schlegl2017unsupervised}, making \ood detection a crucial research problem.

Existing \ood detection methods can be broadly categorized {according to} whether they require finetuning of a  DNN. A series of post-hoc methods \cite{hendrycks2017baseline,liu2020energy,lee2018simple, sun2022out,fang2024kernel,mueller2025mahalanobispp,fang2026kpca,sevillano2026stoodx} {have been proposed with the main idea of}  leveraging the responses from a pretrained model to devise detection scores that capture the \ind-\ood discrepancy. 
Another line of research introduces outliers as surrogate \ood data to finetune the DNN with regularization terms, which effectively promotes  detection performance, known as Outlier Exposure (OE) \cite{hendrycks2019oe}.
OE-based methods advance \ood detection in two main aspects:  {\it outlier construction} and {\it outlier regularization}.
The former improves auxiliary outliers to provide surrogate \ood supervision through boundary-aware mining \cite{chen2021atom,ming2022poem}, ID-outlier mixing \cite{zhang2023mixture}, and informative extrapolation \cite{zhu2023diversified}.
The latter develops regularization objectives to finetune the DNN for enhanced \ind-outlier separation via energy-based margin regularization~\cite{liu2020energy}, gradient regularization of the \ood score function~\cite{sharifi2024gradient}, or feature-based geometric constraints~\cite{wu2025pfs,xia2026vpc}.
Our work falls into the outlier-regularization category.

Early OE methods rely on the output logit space for both outlier generation and regularization \cite{wu2025pfs}.
The logit layer in DNNs primarily encodes coarse-grained knowledge associated with predictive classification probabilities. In contrast, feature spaces of DNNs encode far richer and structured information of data distributions, including hierarchical representations, geometric patterns, and semantics learned across layers. 
Recent feature-based OE methods promote \ind--\ood separation using geometric references derived from classifier weights \cite{wu2025pfs} or predefined structures \cite{xia2026vpc}.
Please find a detailed discussion in Section \ref{sec:oe-ood}.

In this work, we propose \ours, an OE-based \ood detection framework that exploits \ind-\ood discrepancy through optimizing low-dimensional feature subspaces.
We posit that distribution-associated dominant directions of feature variation provide useful structure for distinguishing \ind data from outliers. 
Accordingly, we construct and optimize an \ind-associated subspace and an auxiliary-outlier-associated subspace spanned by learnable orthonormal bases ${\bm U}_{\rm{in}}$ and ${\bm U}_{\rm{aux}}$, respectively. 
As commonly in OE-based detection, auxiliary outliers serve as a surrogate non-\ind reference when the open-world \ood distribution is unavailable during training.
The subspace spanned by ${\bm U}_{\rm {aux}}$ therefore characterizes the structures exposed by these auxiliary training outliers, providing an important reference basis for separating \ind data from \ood data.

In our \ours framework, each subspace is constructed to capture the dominant variations of its corresponding feature distribution while those two subspaces are regularized to remain separated from each other.
To this end, we jointly optimize the subspaces and the backbone, allowing the feature representations to adapt to the fitting and separation objectives. 
We therefore formulate two complementary learning objectives, {\it subspace fitting} and {\it subspace separation}, detailed as follows.

\begin{itemize}
    \item {\textbf {Subspace Fitting}}. {Along with Principal Component Analysis (PCA) \cite{pearson1901liii}, we optimize ${\bm U}_{\rm{in}}$ and ${\bm U}_{\rm{aux}}$ by minimizing the reconstruction error of training samples on ${\bm U}_{\rm{in}}$ and that of outliers on ${\bm U}_{\rm{aux}}$, such that ${\bm U}_{\rm{in}}$ captures intrinsic \ind patterns and ${\bm U}_{\rm{aux}}$ maximally describes the outliers to characterize  the potential \ood.}
    \item {\textbf {Subspace Separation}}.
    The separation between the two subspaces is optimized at both  sample and parameter levels.
    At the sample level, we {deploy the margin-based hinge loss to encourage \ind features with large reconstruction errors on the  subspace ${\bm U}_{\rm {aux}}$, similarly for outlier features on ${\bm U}_{\rm{in}}$.
    This ensures ${\bm U}_{\rm{in}}$ and ${\bm U}_{\rm{aux}}$ to retain minimal information from the opposing  distribution.}
    At the parameter level, an orthogonality regularizer is imposed to disentangle the two subspaces ${\bm U}_{\rm{in}}$ and ${\bm U}_{\rm {aux}}$.
\end{itemize}

\begin{figure}[ht]
    \centering
    \includegraphics[width=0.9\columnwidth]{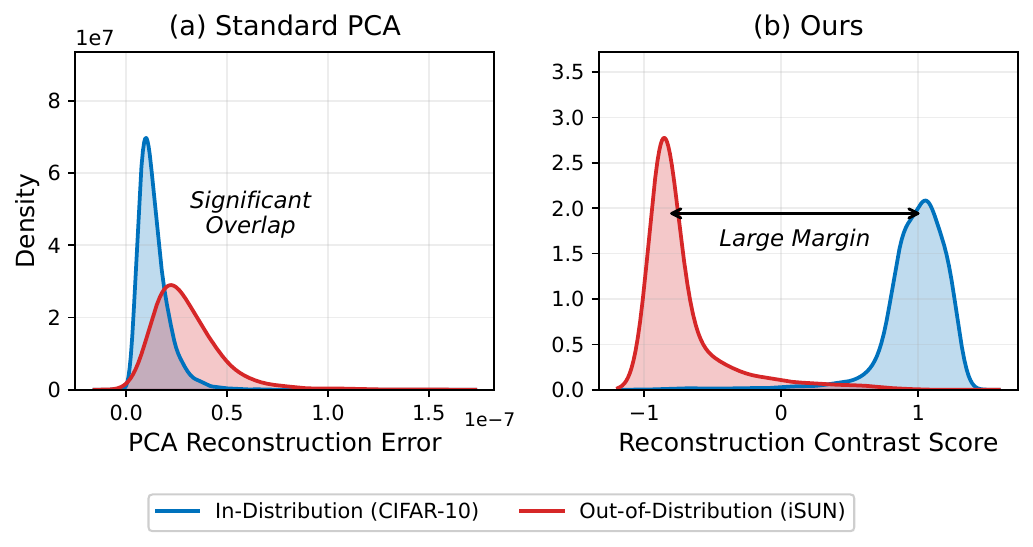} 
    \caption{Experimental exemplification on the effectiveness of \ours, with CIFAR-10 as InD and iSUN as \ood. We interpolate the distribution of \emph{(a)} reconstruction errors from a standard PCA  on the pretrained DNN and \emph{(b)} reconstruction contrast score by \ours. 
    }
    \label{fig:largin_margin}
\end{figure}

Notably in optimization, rather than solving an eigenvalue problem in standard PCA, the orthonormal bases can be optimized on Stiefel manifold \cite{edelman1998geometry} with Riemannian Adam optimizer \cite{becigneul2019riemannian,kochurov2020geoopt}.
Thus, \ours flexibly adapts to the end-to-end  manner under OE-based training framework for \ood detection. 
We further introduce progressive-rank optimization to capture dominant feature variations at low rank before gradually expanding the subspaces to the target rank. This helps reduce interference from poorly estimated directions during early training.

The two optimized subspaces yield contrasting reconstructions of \ind and \ood features.
Specifically, \ind samples are expected to be well reconstructed by ${\bm U}_{\rm{in}}$ and poorly reconstructed by ${\bm U}_{\rm{aux}}$, whereas \ood samples tend to exhibit the opposite behavior.
We use the difference between the two reconstruction errors as an effective detection score, as illustrated in Figure~\ref{fig:largin_margin}.

Our main contributions  are summarized as follows.
\begin{itemize}
    \item We introduce a feature-subspace perspective for OE regularization, revealing that intrinsic low-dimensional structures of \ind and outlier features provide promising discriminative supervision for \ood detection.
    \item We propose \ours, a new framework that learns distribution-specific feature structures with \ind-\ood separability in the  subspaces. By fitting each subspace to its own distribution and suppressing information from the opposite one, their relative reconstruction naturally leads to a discriminative detection score. We introduce progressive-rank optimization to prioritize dominant feature variations early in training.
    \item Extensive experimental results validate the superiority of {\ours} with state-of-the-art performance gains at only slight extra overhead from two small-size matrices.

\end{itemize}

\section{Preliminaries}
\subsection{OE-based \ood Detection}
\label{sec:oe-ood}
In \ood detection, the key is to devise a score function $S(\bm x)$ that reflects the likelihood of a test sample $\bm x$ belonging to the distribution of training data (InD):
\begin{equation}
{\rm category\ of\ } \bm x =
\begin{cases}
\text{InD}, & \text{if } S(\bm x) \ge \tau, \\
\text{\ood}, & \text{otherwise}.
\end{cases}
\end{equation}
The threshold $\tau$ is chosen such that a large proportion of InD data is correctly classified.
Generally, given a target  model $f(\cdot)$, its responses w.r.t. the given sample $\bm x$, such as logits and features, are utilized to devise the scoring function $S(\cdot)$.
We then outline a series of representative \ood detection methods based on Outlier Exposure (OE) \cite{hendrycks2019oe}, which are closely related.

{OE-based} methods introduce outliers as regularization to finetune $f(\cdot)$ with the following learning objective:
\begin{equation}
\label{eq:oe}
\min_{f}\quad
\mathbb{E}_{(\bm{x},y)\sim\mathcal{D}_{\mathrm{tr}}}
\left[\mathcal{L}_{\mathrm{CE}}\bigl(f(\bm{x}),y\bigr)\right]
+
\lambda\,
\mathbb{E}_{\bm{x}'\sim\mathcal{D}_{\mathrm{aux}}}
\left[\mathcal{L}_{\mathrm{OE}}\bigl(f(\bm{x}')\bigr)\right].
\end{equation}
In Eq. \eqref{eq:oe}, $\mathcal{D}_{\rm{tr}}$ and $\mathcal{D}_{\rm{aux}}$ denote the training dataset and the auxiliary outlier dataset, respectively, and $\mathcal{L}_{\rm{CE}}$ denotes the cross-entropy loss. The performance of $f(\cdot)$ on outliers is regularized through an OE regularization loss $\mathcal{L}_{\rm{OE}}$, which is typically defined as the KL divergence between the outlier predictions $f(\bm x')$ and a uniform distribution.
In this way, outliers act as surrogates for \ood data, allowing $f(\cdot)$ to be trained with the regularization term $\mathcal{L}_{\rm{OE}}$ that strengthens the separation between InD and \ood samples.

A series of OE-based methods explore in different ways to {\it generate} diverse outliers.
MixOE \cite{zhang2023mixture} conducts pixel-level interpolation between training samples and outliers; DivOE \cite{zhu2023diversified} introduces adversarial perturbations on outliers; 
DREAM-OOD \cite{du2023dream} leverages diffusion models in the latent space to generate photo-realistic outliers. 
Meanwhile, another line of OE-based research focuses on designing advanced {\it regularization} terms $\mathcal{L}_{\rm{OE}}$ to finetune the DNN.
In \cite{choi2023balanced}, a balanced Energy regularization loss is introduced by minimizing the Energy margins between InD training data and outliers; GReg~\cite{sharifi2024gradient} imposes gradient penalties to incorporate  local information of data for finetuning.

\paragraph{Comparison with related methods}
Among OE regularization methods, approaches in feature space use auxiliary outliers to shape representation geometry.
PFS~\cite{wu2025pfs} imposes separation constraints using the subspace spanned by classifier weights and its orthogonal complement, whereas VPC~\cite{xia2026vpc} uses predefined geometric references.
\ours learns separate subspaces for \ind and auxiliary-outlier features, each fitted to the dominant variations of the corresponding distribution.
On the other hand, subspace structure is also used in post-hoc detection with pretrained backbones.
ViM~\cite{wang2022vim} combines principal-subspace residuals with classification logits, while KPCA~\cite{fang2026kpca} measures reconstruction errors in kernel space.
WDiscOOD~\cite{chen2023wdiscood} combines scores from discriminative and residual subspaces obtained through whitened linear discriminant analysis of \ind features.
These methods derive their subspace references from \ind features while keeping the backbone fixed.
\ours uses auxiliary outliers as a proxy for \ood data and jointly optimizes both subspace bases and the backbone.
This joint learning shapes the representations to preserve distribution-specific structure and promote \ind-outlier separation.

\subsection{PCA Techniques}
PCA \cite{pearson1901liii} has long been a fundamental tool  to analyze high-dimensional data.
It seeks a low-dimensional subspace that preserves the dominant variations of data and has been widely used for dimensionality reduction and feature representation analysis \cite{fang2026kpca}.

Given a centered data/feature vector $\bm z \in \mathbb{R}^{d}$ (i.e., $\bar{\bm z}=0$), PCA seeks a linear projection matrix $\bm U \in \mathbb{R}^{d \times k}$, typically with $k \ll d$, that maps $\bm z$ onto a $k$-dimensional principal subspace. The matrix $\bm U$ contains a set of orthonormal bases spanning the projected subspace, such that  $\bm U^\top \bm U=\bm I_k$, which can also be regarded as residing in the celebrated Stiefel manifold \cite{edelman1998geometry}. PCA can be equivalently optimized through maximizing the projection variation or minimizing the reconstruction error onto the subspace, the latter of which is adopted in our method.

Approaching from the reconstruction-based optimization,  $\bm z$ is projected into the subspace and then mapped, i.e., ${\bm z}_{\rm {rec}} = \bm U \bm U^\top \bm z$.
The discrepancy lost during this process is formally defined as the reconstruction error:
\begin{equation}
\label{eq:reconstruction-error}
e(\bm z; \bm U) = \|\bm z - \bm U \bm U^\top \bm z\|_2^2,
\end{equation}
which quantifies the residual energy of the feature vector lying orthogonal to the learned subspace.
{Small values indicate better reconstructions, i.e., the most intrinsic patterns of $\bm z$ are well retained in this subspace.} Conversely, a large reconstruction error implies that the feature contains significant variations falling outside this subspace. {This inherent property is crucial for motivating the optimization and also the inference in our method as introduced in Section \ref{sec:method}.}

\paragraph{Notation}
DNNs with standard $C$-class classification settings are considered. Let  $\mathcal{X} \subset \mathbb{R}^m$ denote the input space and $\mathcal{Y} \subset \{1,2,\ldots,C\}$ denote the label space.
We denote the InD training dataset and auxiliary outlier dataset by $\mathcal D_{\mathrm{tr}}$ and $\mathcal D_{\mathrm{aux}}$,  respectively.
A DNN is defined as $f(\cdot) = h(g(\cdot))$, where $g: \mathbb{R}^{m} \rightarrow \mathbb{R}^{d}$ denotes the backbone encoder and $h: \mathbb{R}^{d} \rightarrow \mathbb{R}^{C}$ denotes the classifier head. For an input sample $\bm x$, we denote its penultimate-layer feature as $\bm z = g(\bm x) \in \mathbb{R}^{d}$.

\section{Methodology}\label{sec:method}
\begin{figure*}[t]
    \centering
    \includegraphics[width=\textwidth]{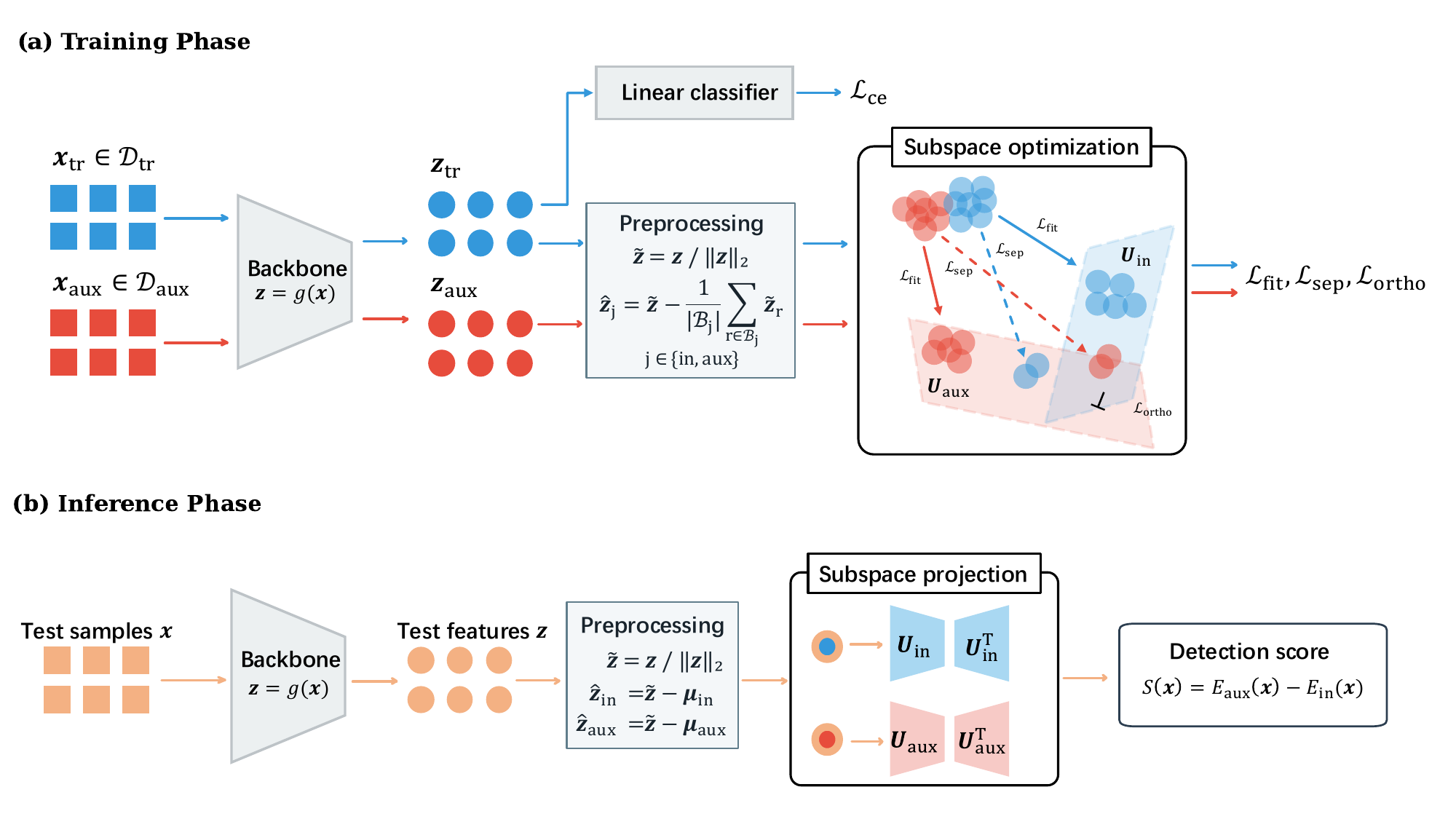} 
    \caption{Illustration of our  \ours for OE-based \ood detection. 
    \emph{In training}, \ours seeks  separation between \ind and outliers  through two subspaces spanned by $\bm U_{\rm {in}}$ and $\bm U_{\rm {aux}}$, which are optimized not only to capture maximal information of \ind data and \ood data respectively (i.e., $\mathcal{L}_{\text{fit}}$), but also to encode minimal information  from the opposing data distribution (i.e.,  $\mathcal{L}_{\text{sep}}$), together with an orthogonality term (i.e., $\mathcal{L}_{\text{ortho}}$) to further refine the two subspaces. {\emph{At inference}, the difference of reconstruction errors in the learned two subspaces is naturally utilized as detection scores.}}
    \label{fig:framework}
\end{figure*}

In this section, we present \ours, an end-to-end feature-subspace learning framework for OE-based OoD detection.
As illustrated in Figure \ref{fig:framework},  our framework  characterizes the intrinsic patterns of \ind and \ood samples through the optimization with two feature subspaces.
We first elaborate the subspace modeling, and then introduce our optimization scheme.

\subsection{Subspace Modeling}
\label{sec:method:subspace-modeling}
\paragraph{Feature Preprocessing}
Following previous feature-space modeling methods \cite{fang2024kernel,wu2025pfs}, we apply centered $\ell_2$ normalization before learning the subspaces. Each feature is first normalized as $\widetilde{\bm z}_i=\bm z_i/\lVert\bm z_i\rVert_2$.
For each distribution $j\in\{\mathrm{in},\mathrm{aux}\}$, we compute the mean of its normalized features in the current mini-batch $\mathcal B_j$: $\overline{\bm z}_j = \frac{1}{|\mathcal B_j|} \sum_{i\in\mathcal B_j}\widetilde{\bm z}_i.$
During training, this batch mean is used for centering and to update a distribution-specific exponential moving average (EMA) center $\bm\mu_j$. Each EMA center is initialized from the first batch mean
and subsequently updated as $\bm\mu_j \leftarrow 0.99\,\bm\mu_j+0.01\,\overline{\bm z}_j.$
The centered representation used by subspace $j$ is
\begin{equation}
\widehat{\bm z}_{i,j}
=
\begin{cases}
\widetilde{\bm z}_i-\overline{\bm z}_j,
& \text{during training},\\
\widetilde{\bm z}_i-\bm\mu_j,
& \text{during inference}.
\end{cases}
\end{equation}
The center is always selected according to the subspace performing reconstruction, including when evaluating features from the opposing distribution.
During inference, the EMA centers remain fixed, so centering does not depend on the composition of the test batch.

The two preprocessing operations serve complementary purposes. The $\ell_2$ normalization removes differences in feature magnitude, making projection and reconstruction-error measurements depend on the normalized representations rather than their original scales.
Mean centering follows the principle used in PCA, removing the distribution mean before identifying principal directions of variation.
Each subspace therefore models deviations around its distribution-specific center.
When a feature is evaluated against the opposing distribution's subspace, using that subspace's center measures reconstruction relative to the corresponding distribution.

\paragraph{Feature Subspace} 
We construct two subspaces to model the dominant variations of \ind features and \ood features, respectively.
Since the complete open-world \ood distribution is unavailable during training, outliers are used to learn a proxy \ood subspace that captures non-\ind structures observed during training.
Along with the spirits in PCA, we learn each subspace by minimizing reconstruction errors for its associated distribution, while encouraging separation between the two subspaces.

We denote that the subspaces are spanned by two orthonormal bases ${\bm U}_{\rm{in}}\in\mathbb{R}^{d\times k}$ and ${\bm U}_{\rm{aux}}\in\mathbb{R}^{d\times k}$, naturally residing on the Stiefel manifold $\text{St}(d,k)$, i.e., $\bm U_{\rm {in}}^\top \bm U_{\rm {in}} = \bm I_k$ and $\bm U_{\rm {aux}}^\top \bm U_{\rm {aux}} = \bm I_k$.  
For an input $\bm x$ with feature $\bm z=g(\bm x)$,
let $\widehat{\bm z}_j(\bm x)$ denote its normalized feature
centered using the center associated with subspace $j$,
as defined in Feature Preprocessing.
Following Eq.~\eqref{eq:reconstruction-error}, we define
\begin{equation}
\label{eq:preprocessed-error}
E_j(\bm x)
:=
\left\|
\widehat{\bm z}_j(\bm x)
-
\bm U_j\bm U_j^\top\widehat{\bm z}_j(\bm x)
\right\|_2^2,
\qquad j\in\{\mathrm{in},\mathrm{aux}\}.
\end{equation}
For each subspace $j$, $E_j(\bm x)$ measures the squared residual that remains after orthogonal projection onto the corresponding subspace. A smaller $E_j(\bm x)$ indicates that the subspace $j$ better explains the feature relative to its distribution-specific center.

\paragraph{Detection Score}

Given the learned subspaces, we define the detection score for a test sample $\bm x$ as
\begin{equation}
\label{eq:det-score}
S(\bm x)
=
E_{\mathrm{aux}}(\bm x)
-
E_{\mathrm{in}}(\bm x).
\end{equation}
This score compares how well the two subspaces reconstruct the feature relative to their respective centers. The learning objective encourages \ind features to have lower reconstruction errors in the \ind subspace than in the proxy \ood subspace, yielding higher scores. Lower scores indicate greater relative compatibility with the proxy \ood subspace and are used to identify \ood samples.

\subsection{Subspace Optimization}
\label{sec:method:subspace-opt}

We now elaborate our optimization strategy to obtain the aforementioned desired feature subspaces ${\bm U}_{\rm{in}}$ and ${\bm U}_{\rm {aux}}$.
The optimization enforces two complementary principles: each feature distribution is well aligned with its corresponding subspace, while the two learned subspaces are explicitly separated from each other.

\paragraph{Subspace Fitting}
The two feature subspaces ${\bm U}_{\rm{in}}$ and ${\bm U}_{\rm {aux}}$ are primarily designed to capture the intrinsic patterns of their respective data distributions, {similar to the principal components in PCA}. 
We establish the objective of minimizing the reconstruction errors on the \ind training samples ${\cal D}_{\rm tr}=\{(\bm x^{(i)}_{\rm tr},y_{\rm tr}^{(i)})\}_{i=1}^{N_{\rm tr}}$ onto ${\bm U}_{\rm{in}}$ and the outliers ${\cal D}_{\rm aux}=\{\bm x^{(i)}_{\rm aux}\}_{i=1}^{N_{\rm aux}}$ onto ${\bm U}_{\rm {aux}}$, respectively.
This design enables each subspace to encode the information most relevant to its own data. To be specific, this loss is given by:
\begin{equation}
\label{eq:loss:fitting}
{\cal L}_{\rm fit}
=
\sum_{i=1}^{N_{\rm tr}}
E_{\mathrm{in}}(\bm x_{\rm tr}^{(i)})
+
\sum_{i=1}^{N_{\rm aux}}
E_{\mathrm{aux}}(\bm x_{\rm aux}^{(i)}).
\end{equation}
\paragraph{Subspace Separation}
For \ood detection, we encourage each subspace to distinguish features from its corresponding distribution from those of the opposing distribution. Specifically, we penalize reconstruction errors below a prescribed margin when features are projected onto the opposing subspace, yielding the following hinge loss:
\begin{equation}
\label{eq:loss:separation}
{\cal L}_{\rm sep}
=
\sum_{i=1}^{N_{\rm tr}}
\max\!\left(
0,\,
m-E_{\mathrm{aux}}(\bm x_{\rm tr}^{(i)})
\right)
+
\sum_{i=1}^{N_{\rm aux}}
\max\!\left(
0,\,
m-E_{\mathrm{in}}(\bm x_{\rm aux}^{(i)})
\right),
\end{equation}
where $m$ is the margin in hinge loss.
Minimizing ${\cal L}_{\rm sep}$ encourages cross-subspace reconstruction errors to reach or exceed $m$, with no further penalty once the margin is satisfied. Together with ${\cal L}_{\rm fit}$, this objective encourages each feature to be reconstructed more accurately by its corresponding subspace. We further penalize overlap between the two subspaces through ${\cal L}_{\rm ortho}=\lVert \bm U_{\rm{in}}^{\top}{\bm U}_{\rm{aux}}\rVert_F^2$. This parameter-level regularizer complements sample-level separation objective by discouraging shared directions between the two learned bases.

\begin{figure*}[t]
    \centering
    \includegraphics[width=\linewidth]{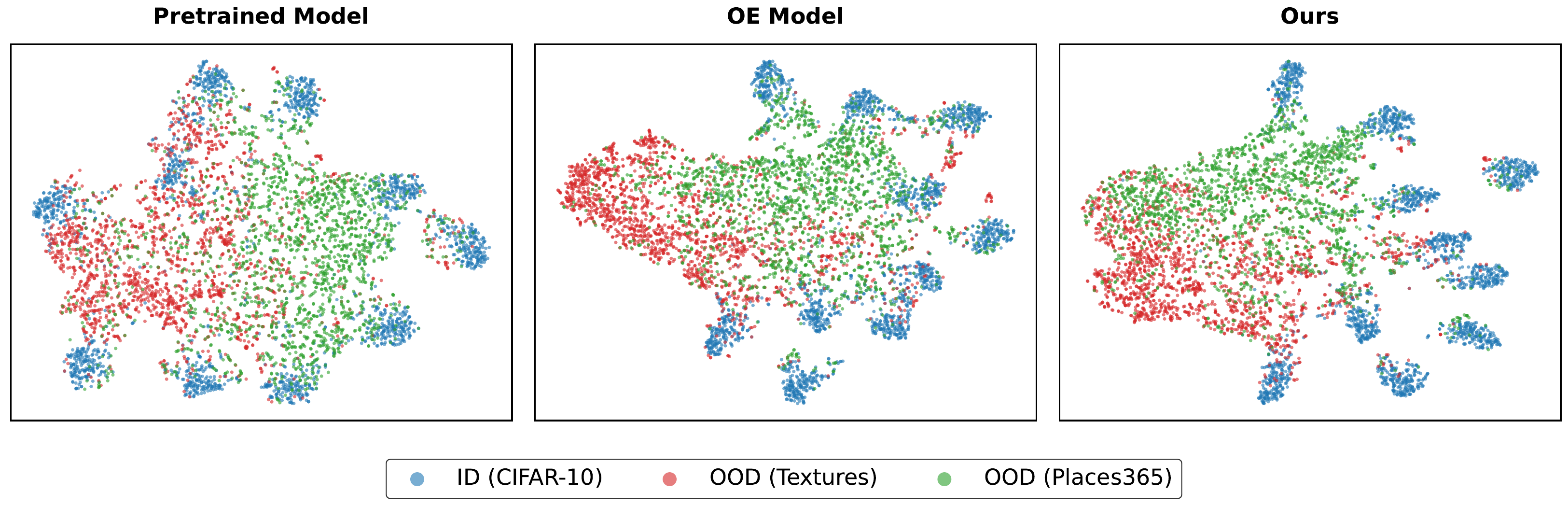}
    \caption{T-SNE visualization of the learned features for InD and OoD data. \ours  achieves a  clearer separation between InD (blue) and \ood (red and green) features compared to the Pretrained Model and the OE Model. By regularizing the network with subspace loss term, the network is able to output more distinguishable InD and \ood features. This distinct feature separation inherently facilitates the subsequent \ood detection based on reconstruction errors.}
    \label{fig:T-SNE}
\end{figure*}

\paragraph{Joint Optimization}
With the aforementioned schemes for subspace fitting and subspace separation, we formulate the overall optimization objective for our subspace-based OE regularization as follows:
\begin{equation}
\label{eq:overall-opt}
\begin{aligned}
\min_{f(\cdot),{\bm U}_{\rm{in}},{\bm U}_{\rm{aux}}}
\quad&
{\cal L}_{\rm CE}
+\gamma{\cal L}_{\rm sub},
\end{aligned}
\end{equation}
where ${\cal L}_{\rm sub} = {\cal L}_{\rm fit} +{\cal L}_{\rm sep}+\beta{\cal L}_{\rm ortho}$,
${\cal L}_{\rm CE}$ is the standard cross-entropy loss, and coefficients $\gamma$ and $\beta$ penalize the corresponding loss terms. 
During training, ${\cal L}_{\rm sub}$ jointly updates the two subspace bases and the backbone $g$, allowing the feature representations to adapt to the subspace fitting and separation objectives. Meanwhile, ${\cal L}_{\rm CE}$ encourages the retention of class-discriminative information. This joint optimization promotes separation between \ind and \ood features within the learned representation space. The feature visualization in Figure~\ref{fig:T-SNE} further illustrates the resulting separation between \ind and \ood samples.

The  parameters in optimization include the pretrained DNN $f(\cdot)$ and the two orthonormal matrices ${\bm U}_{\rm{in}}$ and ${\bm U}_{\rm {aux}}$.
We adopt standard stochastic gradient descent optimizer to update the network parameters in $f(\cdot)$, in line with the settings in existing OE-based methods, while using the Riemannian Adam optimizer \cite{becigneul2019riemannian,kochurov2020geoopt} to update the two matrices ${\bm U}_{\rm{in}}$ and ${\bm U}_{\rm {aux}}$ constrained on the Stiefel manifold, i.e., $\bm U_{\rm {in}}^\top \bm U_{\rm {in}} = \bm I_k$ and $\bm U_{\rm {aux}}^\top \bm U_{\rm {aux}} = \bm I_k$.

\subsection{Progressive-Rank Optimization}
During optimization, the subspace bases adapt to the evolving feature representations produced by the backbone.
We therefore introduce subspace capacity progressively, allowing a smaller set of directions to be optimized before additional directions are incorporated.

Let $k$ denote the target rank and $r^{(s)}$ the shared active rank of the two subspaces at stage $s$.
Starting from $r^{(0)}=\max\{1,\lceil k/2\rceil\}$, we expand both subspaces every $K$ optimization steps according to
\begin{equation}
r^{(s+1)}
=
\min\left\{
k,\,
r^{(s)}
+
\max\left\{
1,\,
\left\lceil
\frac{k-r^{(s)}}{2}
\right\rceil
\right\}
\right\}.
\label{eq:progressive_half_remaining}
\end{equation}
This schedule gradually introduces the remaining directions until the target rank is reached.

At each expansion, we preserve the learned subspace and initialize additional directions in its orthogonal complement.
The expanded basis remains orthonormal, and all directions continue to be jointly optimized.
This procedure increases the representation capacity while retaining the subspace learned in preceding stages.
Training uses the current-rank bases, whereas inference uses the final rank-$k$ bases.
The initialization, expansion procedure, optimization settings and complete algorithm are detailed in Appendix A.2.

\section{Experiments}
In this section, we present extensive numerical experiments to evaluate the proposed \ours  with comparisons to a series of classic and strong baselines covering post-hoc and OE-based methods {in Section~\ref{sec:main_comparisons}}. We also conduct multiple ablation studies on  the roles of our optimized subspaces and the  mechanisms fundamental to accurate  detections {in Section.~\ref{sec:ablation}}.

\subsection{Experimental Setups}
The main protocol in our experiments mainly follows
\cite{zhu2023diversified}, covering data, models,
and training strategies.

\noindent \textbf{Datasets.}
With the established protocols in \cite{zhu2023diversified,hendrycks2019oe, wang2023implicit}, we experiment on CIFAR-10 \cite{krizhevsky2009learning} and ImageNet-1K \cite{deng2009imagenet} benchmarks. 
For CIFAR-10 as \ind, we assess \ood detection performance using  Textures \cite{cimpoi2014describing}, Places365 \cite{zhou2016places}, iSUN \cite{xu2015turkergaze}, LSUN\_Crop \cite{yu2015lsun}, and LSUN\_Resize \cite{yu2015lsun} as \ood datasets, as typically adopted in the community. Tiny\_ImageNet \cite{le2015tiny} is utilized as the auxiliary surrogate \ood dataset during OE-based training. 
For \textit{ImageNet}, we use ImageNet-1K as the InD dataset and ImageNet-21K\_p \cite{ridnik2021imagenet} as the surrogate \ood data. The \ood data for ImageNet are iNaturalist \cite{van2018inaturalist}, SUN \cite{xiao2010sun}, Places \cite{zhou2016places}, and Textures \cite{cimpoi2014describing}.

\noindent \textbf{Implementation.} We employ WideResNet40-2 \cite{zagoruyko2016wide} and ResNet50 \cite{he2016deep} as backbones for CIFAR and ImageNet, respectively. 
Following previous OE methods \cite{zhu2023diversified}, we finetune for 10 epochs on CIFAR-10 and 4 epochs on ImageNet, respectively. For \ours, the backbone is optimized using SGD with cosine annealing, while the matrices $\bm U_{\rm {in}}$ and $\bm U_{\rm {aux}}$ are learned with the Riemannian Adam optimizer \cite{becigneul2019riemannian,kochurov2020geoopt}.

\noindent \textbf{Baselines.}
We compare our proposed \ours with two main categories of  classic and recent strong baselines: 
\emph{(i)} \textit{Post-hoc methods}, including MSP \cite{hendrycks2017baseline}, ODIN \cite{liang2018odin}, Mahalanobis \cite{lee2018simple}, Energy \cite{liu2020energy}, ASH \cite{djurisic2023ash}, PCA \cite{guan2023revisit}, KPCA (CoP and CoRP) \cite{fang2026kpca}, and Mahalanobis++ \cite{mueller2025mahalanobispp}; \emph{(ii)} \textit{OE-based methods}, including standard OE \cite{hendrycks2019oe}, EnergyOE \cite{liu2020energy}, POEM \cite{ming2022poem}, MixOE \cite{zhang2023mixture}, DivOE \cite{zhu2023diversified}, GReg \cite{sharifi2024gradient}, PFS \cite{wu2025pfs} and VPC\cite{xia2026vpc}. Three typical metrics are employed for evaluating detection results: \emph{(i)} \textit{FPR95}, the false positive rate of \ood samples at a 95\% true positive rate for \ind test data; \emph{(ii)} \textit{AUROC}, the area under the receiver operating characteristic curve; and \emph{(iii)} \textit{ID-ACC}, the classification accuracy on InD test data.
Further experimental details are provided in Appendix A.1 and Appendix A.3.

\begin{table}[!t]
\centering
\resulttable
\footnotesize
\renewcommand{\arraystretch}{1.04}
\begin{tabularx}{\linewidth}{
    >{\hsize=1.30\hsize\linewidth=\hsize\centering\arraybackslash}X
    >{\hsize=0.90\hsize\linewidth=\hsize\centering\arraybackslash}X
    >{\hsize=0.90\hsize\linewidth=\hsize\centering\arraybackslash}X
    >{\hsize=0.90\hsize\linewidth=\hsize\centering\arraybackslash}X
}
\toprule

\textbf{Method}
& \textbf{FPR95}$\downarrow$
& \textbf{AUROC}$\uparrow$
& \textbf{ID-ACC}$\uparrow$ \\

\midrule
\multicolumn{4}{l}{\textit{Post-hoc methods}} \\

MSP
& 52.21 & 90.61 & 94.84 \\

ODIN
& 32.84 & 90.67 & 94.84 \\

Mahalanobis
& 36.27 & 92.69 & 94.84 \\

Energy
& 32.82 & 91.99 & 94.84 \\

ASH
& 33.01 & 92.11 & 94.84 \\

PCA
& 69.84 & 74.82 & 94.84 \\

KPCA (CoP)
& 40.22 & 91.28 & 94.84 \\

KPCA (CoRP)
& 31.82 & 93.80 & 94.84 \\

Mahalanobis++
& 19.56 & 95.90 & 94.84 \\

\midrule
\multicolumn{4}{l}{\textit{OE-based methods}} \\

OE
& $\stat{13.76}{0.23}$
& $\stat{97.53}{0.03}$
& $\stat{94.51}{0.06}$ \\

EnergyOE
& $\stat{18.15}{0.32}$
& $\stat{90.90}{0.18}$
& $\stat{94.51}{0.01}$ \\

POEM
& $\stat{17.81}{0.78}$
& $\stat{90.96}{0.37}$
& $\stat{94.90}{0.04}$ \\

MixOE
& $\stat{13.57}{0.25}$
& $\stat{97.47}{0.07}$
& $\stat{\mathbf{94.98}}{0.02}$ \\

DivOE
& $\stat{11.66}{0.04}$
& $\stat{97.82}{0.04}$
& $\stat{94.66}{0.12}$ \\

GReg
& $\stat{12.69}{1.78}$
& $\stat{97.03}{0.51}$
& $\stat{94.05}{0.10}$ \\

PFS
& $\stat{8.95}{0.40}$
& $\stat{95.43}{0.13}$
& $\stat{94.36}{0.11}$ \\

VPC
& $\stat{15.12}{0.23}$
& $\stat{96.51}{0.11}$
& $\stat{94.06}{0.05}$ \\

\midrule
\resultrow
\textbf{\ours{} (ours)}
& $\stat{\mathbf{4.36}}{0.76}$
& $\stat{\mathbf{98.68}}{0.24}$
& $\stat{94.65}{0.06}$ \\

\bottomrule
\end{tabularx}
\captionsetup{font=small,skip=6pt}
\caption{
OOD detection results on CIFAR-10 (\%).
Results are reported as mean $\pm$ standard
deviation over three trials when available.
}
\label{tab:main_results_cifar}
\end{table}

\begin{table}[!t]
\centering
\resulttable
\footnotesize
\renewcommand{\arraystretch}{1.04}
\begin{tabularx}{\linewidth}{
    >{\hsize=1.30\hsize\linewidth=\hsize\centering\arraybackslash}X
    >{\hsize=0.90\hsize\linewidth=\hsize\centering\arraybackslash}X
    >{\hsize=0.90\hsize\linewidth=\hsize\centering\arraybackslash}X
    >{\hsize=0.90\hsize\linewidth=\hsize\centering\arraybackslash}X
}
\toprule

\textbf{Method}
& \textbf{FPR95}$\downarrow$
& \textbf{AUROC}$\uparrow$
& \textbf{ID-ACC}$\uparrow$ \\

\midrule
\multicolumn{4}{l}{\textit{Post-hoc methods}} \\

MSP
& 75.23 & 76.67 & 68.72 \\

ODIN
& 67.61 & 80.36 & 68.72 \\

Mahalanobis
& 83.41 & 58.47 & 68.72 \\

Energy
& 71.44 & 79.52 & 68.72 \\

ASH
& 73.51 & 77.03 & 68.72 \\

PCA
& 84.36 & 57.44 & 68.72 \\

KPCA (CoP)
& 58.40 & 83.86 & 68.72 \\

KPCA (CoRP)
& 48.60 & 87.53 & 68.72 \\

Mahalanobis++
& 49.54 & 88.14 & 68.72 \\

\midrule
\multicolumn{4}{l}{\textit{OE-based methods}} \\

OE
& $\stat{61.94}{0.03}$
& $\stat{81.58}{0.01}$
& $\stat{75.90}{0.00}$ \\

EnergyOE
& $\stat{66.94}{0.01}$
& $\stat{79.93}{0.02}$
& $\stat{74.19}{0.08}$ \\

POEM
& $\stat{67.81}{0.44}$
& $\stat{79.36}{0.24}$
& $\stat{74.21}{0.03}$ \\

MixOE
& $\stat{70.12}{0.26}$
& $\stat{78.85}{0.05}$
& $\stat{76.18}{0.01}$ \\

DivOE
& $\stat{60.12}{0.11}$
& $\stat{81.96}{0.00}$
& $\stat{75.73}{0.02}$ \\

GReg
& $\stat{64.11}{4.27}$
& $\stat{83.42}{1.48}$
& $\stat{73.02}{0.51}$ \\

PFS
& $\stat{60.19}{0.21}$
& $\stat{83.27}{0.10}$
& $\stat{\mathbf{76.35}}{0.01}$ \\

VPC
& $56.50$
& $84.70$
& $76.11$ \\

\midrule
\resultrow
\textbf{\ours{} (ours)}
& $\stat{\mathbf{36.32}}{2.04}$
& $\stat{\mathbf{91.06}}{0.20}$
& $\stat{75.69}{0.07}$ \\

\bottomrule
\end{tabularx}
\captionsetup{font=small,skip=6pt}
\caption{
OOD detection results on ImageNet-1K (\%).
Results are reported as mean $\pm$ standard
deviation over three trials when available.
}
\label{tab:main_results_imagenet}
\end{table}

\subsection{Main Comparisons}
\label{sec:main_comparisons}
The main comparisons of our \ours with representative \ood detection methods on the CIFAR-10 and ImageNet-1K benchmarks are presented in Table~\ref{tab:main_results_cifar} and Table~\ref{tab:main_results_imagenet}. The reported results are averaged across multiple \ood test datasets, while the results on individual datasets are provided in Appendix B.
In Tables~\ref{tab:main_results_cifar} and~\ref{tab:main_results_imagenet}, post-hoc results are checkpoint-specific point estimates, as each setting is evaluated on a single fixed checkpoint without independent repeated trials. OE results are reported as mean $\pm$ standard deviation over three trials when available; published results lacking these statistics are retained as point estimates. For VPC~\cite{xia2026vpc}, we reproduce the CIFAR-10 results and use the published ImageNet-1K results.
The published ImageNet-1K setting uses five fine-tuning epochs, compared with four epochs in our experiments, and does not report standard deviations.

As illustrated, {\ours} significantly outperforms the compared baselines in both FPR95 and AUROC by a substantial margin, while maintaining competitive \ind classification accuracy.  
This joint strength on \ood robustness and \ind performance is challenging for the baseline methods.
For instance, our method yields substantial improvements over the very recent VPC \cite{xia2026vpc} on ImageNet with FPR95  reduced by 20.18 percentage points  and AUROC boosted by 6.36 percentage points, despite VPC’s {readily strong performance with 56.50\% on FPR95  and 84.70\% on AUROC.}
Such improvements validate our effectiveness  through  subspaces optimized for  \ind and \ood. {This result demonstrates great potential of this new perspective of proceeding outlier regularization on feature principal spaces, rather than the output logit space.}

To evaluate \ours under different degrees of distributional proximity, we further evaluate the tasks on far-\ood and near-\ood with the ImageNet-1K benchmark and compare with strong baselines, following protocols in \cite{wu2025pfs}.
Table \ref{tab:far_near_ood_results} shows that \ours achieves competitive far-\ood performance, with FPR95 of 31.46\% and  AUROC of 93.60\%, although it  trails  Mahalanobis++ by 3.53 percentage points and 1.18 percentage points, respectively. 
On near-\ood datasets, \ours significantly reduces FPR95 by 22.07 percentage points and improves AUROC by 6.21 percentage points  upon the strongest  VPC. These results demonstrate the particular effectiveness of \ours in distinguishing \ood samples lying close to  \ind, which is commonly considered as more challenging settings for \ood detection.

\begin{table}[!htbp]
\centering
{
\small
\setlength{\tabcolsep}{2.5pt}

\begin{tabularx}{\columnwidth}{
    @{}
    >{\centering\arraybackslash}X
    >{\centering\arraybackslash}p{0.16\columnwidth}
    >{\centering\arraybackslash}p{0.16\columnwidth}
    >{\centering\arraybackslash}p{0.16\columnwidth}
    >{\centering\arraybackslash}p{0.16\columnwidth}
    @{}
}
\toprule

\multirow{2}{*}{\textbf{Method}}
& \multicolumn{2}{c}{\textbf{Far-\ood}}
& \multicolumn{2}{c}{\textbf{Near-\ood}} \\

\cmidrule(lr){2-3}
\cmidrule(lr){4-5}

& \shortstack{FPR95$\downarrow$}
& \shortstack{AUROC$\uparrow$}
& \shortstack{FPR95$\downarrow$}
& \shortstack{AUROC$\uparrow$} \\

\midrule

KPCA (CoRP)
& 29.66
& 93.75
& 67.85
& 81.33 \\

Mahalanobis++
& \textbf{27.93}
& \textbf{94.78}
& 71.16
& 81.51 \\

PFS
& 52.23
& 85.62
& 68.16
& 80.92 \\

VPC
& 49.75
& 87.10
& 63.25
& 82.30 \\

\textbf{\ours (ours)}
& 31.46
& 93.60
& \textbf{41.18}
& \textbf{88.51} \\

\bottomrule
\end{tabularx}
}
\caption{
\ood detection results on the detection tasks of far-\ood and near-\ood. 
}
\label{tab:far_near_ood_results}
\end{table}

\paragraph{Computation Cost} 
\label{sec:efficiency}
Our method additionally optimizes two basis matrices, $\bm U_{\rm {in}}\in \mathbb{R}^{d\times k}$ and $\bm U_{\rm {aux}}\in \mathbb{R}^{d\times k}$, for subspace projections. Together, these matrices require storing $2dk$ coefficients at rank $k$. We compare the computational cost of \ours with that of standard OE using WideResNet-40-2 on CIFAR-10, reporting trainable parameter counts and mean fine-tuning time per epoch. As shown in Table~\ref{tab:computational_cost}, \ours achieves lower FPR95 and higher AUROC, with an 18.4\% increase in per-epoch fine-tuning time and a 0.71\% increase in trainable parameters. These results demonstrate a favorable trade-off between OOD detection performance and the additional training time and parameter overhead introduced by \ours.

\begin{table}[!htbp]
\centering
\resulttable
\begin{tabularx}{\linewidth}{*{5}{>{\centering\arraybackslash}X}}
\toprule
\textbf{Method}
& \textbf{Parameters}
& \shortstack{\textbf{Time}}
& \textbf{FPR95}$\downarrow$
& \textbf{AUROC}$\uparrow$ \\
\midrule
OE
& 2.244M
& $\stat{14.10}{0.30}$
& $\stat{13.76}{0.23}$
& $\stat{97.53}{0.03}$ \\
\resultrow
\ours
& 2.260M
& $\stat{16.70}{1.19}$
& $\stat{4.36}{0.76}$
& $\stat{98.68}{0.24}$ \\
\bottomrule
\end{tabularx}
\caption{Training cost and detection performance on CIFAR-10 with WideResNet-40-2. RecOE parameters are counted at the target rank. Time (s/epoch) and detection metrics are mean $\pm$ standard deviation.}
\label{tab:computational_cost}
\end{table}

\subsection{Ablation Study}
\label{sec:ablation}
To better understand the mechanisms underlying the improved detection performance of \ours, we ablate individual loss terms on CIFAR-10 and evaluate single-subspace variants on ImageNet-1K. We further evaluate the progressive-rank optimization strategy on both CIFAR-10 and ImageNet-1K to assess its effectiveness in improving OOD detection performance.

\paragraph{Individual loss terms}
We conduct a leave-one-out ablation study on the {losses regarding optimizing our subspaces, i.e.,} 
$\mathcal{L}_{\text{sub}}=\mathcal{L}_{\text{fit}}+\mathcal{L}_{\text{sep}}+\beta\mathcal{L}_{\text{ortho}}$, with results reported in Table~\ref{tab:ablation_components}.
The results show that removing either $\mathcal{L}_{\mathrm{fit}}$ or $\mathcal{L}_{\mathrm{sep}}$ leads to a substantial degradation in \ood detection performance, demonstrating that both strategies are essential to our subspace learning framework. Without $\mathcal{L}_{\mathrm{fit}}$, the learned subspaces fail to adequately capture the dominant variations of their corresponding feature distributions. In contrast, removing $\mathcal{L}_{\mathrm{sep}}$ prevents the model from explicitly distinguishing each distribution from the opposing subspace, 
leading to increased overlap between the two learned subspaces.
The comparatively smaller performance drop after removing $\mathcal{L}_{\mathrm{ortho}}$ indicates that $\mathcal{L}_{\mathrm{ortho}}$ plays a complementary role by reducing the overlap between the two subspaces.

\begin{table}[!htbp]
\centering

\setlength{\tabcolsep}{3.2pt}
\begin{tabular*}{\columnwidth}{
    @{\extracolsep{\fill}}
    c c c
    @{\hspace{1.2em}}
    c c
    @{}
}
\toprule
\multicolumn{3}{c}{\textbf{Loss Terms}}
&
\multicolumn{2}{c}{\textbf{\ood Detection Results}} \\
\cmidrule(lr){1-3}
\cmidrule(lr){4-5}

$\mathcal{L}_{\mathrm{fit}}$
& $\mathcal{L}_{\mathrm{sep}}$
& $\mathcal{L}_{\mathrm{ortho}}$
& FPR95$\downarrow$
& AUROC$\uparrow$ \\
\midrule

--         & \checkmark & \checkmark
& 14.89 & 95.83 \\

\checkmark & --         & \checkmark
& 29.19 & 94.57 \\

\checkmark & \checkmark & --
& 4.77 & 98.40 \\

\checkmark & \checkmark & \checkmark
& \textbf{4.36} & \textbf{98.68} \\

\bottomrule
\end{tabular*}

\caption{
Leave-one-out ablation of the subspace loss terms on CIFAR-10 (\%). 
}
\label{tab:ablation_components}
\end{table}

\paragraph{Single Subspace}
We evaluate the dual-subspace design on ImageNet-1K by varying the subspaces used during training and testing, with all other experimental settings unchanged.
We first compare \ours with the variants trained and tested using a single subspace.
As shown in Table~\ref{tab:subspace_ablation}, \ours outperformed both variants in FPR95 and AUROC.
Compared with the stronger InD-only variant, \ours reduces FPR95 by 1.86 percentage points and increases AUROC by 0.48 percentage points.
The results in this ablation demonstrate the  indispensability of our two subspace construction and its optimization, in which the contrast reconstruction between subspaces ${\bm U}_{\rm{in}}$ and ${\bm U}_{\rm {aux}}$ effectively yields discriminative  separation in \ood detection.

\begin{table}[!htbp]
\centering
\resulttable
\begin{tabularx}{\linewidth}{
    >{\centering\arraybackslash}X
    >{\centering\arraybackslash}X
    >{\centering\arraybackslash}p{0.14\linewidth}
    >{\centering\arraybackslash}p{0.14\linewidth}
}
\toprule
\textbf{Training} & \textbf{Testing} & \textbf{FPR95}$\downarrow$ & \textbf{AUROC}$\uparrow$ \\
\midrule
InD-only & InD-only & $38.18$ & $ 90.58 $ \\
Auxiliary-only & Auxiliary-only & $76.61$ & $70.06$ \\
\midrule
Dual-subspace & InD-only & $44.24$ & $88.70$ \\
Dual-subspace & Auxiliary-only & $39.05$ & $88.74$ \\
\resultrow
Dual-subspace & Dual-subspace (\ours) & $\mathbf{36.32}$ & $\mathbf{91.06}$ \\
\bottomrule
\end{tabularx}
\caption{Subspace ablation on ImageNet-1K (\%). Training and testing columns specify the subspaces used at each stage.}
\label{tab:subspace_ablation}
\end{table}

To isolate the role of reconstruction contrast at inference, we retain the trained dual-subspace model and use either subspace alone at test time. Both variants perform worse than contrast scoring, indicating that an individual reconstruction error does not fully exploit the information available in the learned subspace pair. By contrasting reconstruction errors between ${\bm U}_{\rm{in}}$ and ${\bm U}_{\rm {aux}}$, \ours evaluates each sample’s relative compatibility with the two distributions. This comparison supports the effectiveness of jointly using both subspaces at inference for \ood detection.

\paragraph{Progressive-Rank Optimization.}
We compare the progressive-rank strategy with a fixed-rank variant that initializes and optimizes all target-rank directions from the beginning.
Both configurations use the same final subspace rank.
As shown in Table~\ref{tab:ablation_progressive_rank}, progressive-rank optimization achieved modest improvements in both FPR95 and AUROC on the two benchmarks, indicating that directly optimizing the full-rank subspaces is less effective. By first learning compact subspaces and then gradually introducing additional directions, progressive-rank optimization prioritizes dominant feature variations and reduces the interference caused by poorly estimated directions during early training. These results validate the effectiveness of the proposed progressive-rank pipeline in facilitating stable and discriminative subspace learning.

\begin{table}[!htbp]
\centering

\small
\setlength{\tabcolsep}{2pt}
\newcolumntype{Y}{>{\centering\arraybackslash}X}
\begin{tabularx}{\columnwidth}{
    @{}
    >{\raggedright\arraybackslash}p{0.30\columnwidth}
    *{4}{Y}
    @{}
}
\toprule
\multirow{2}{*}{\textbf{Method}}
& \multicolumn{2}{c}{\textbf{CIFAR-10}}
& \multicolumn{2}{c}{\textbf{ImageNet-1K}} \\
\cmidrule(lr){2-3}
\cmidrule(lr){4-5}
& \textbf{FPR95}$\downarrow$
& \textbf{AUROC}$\uparrow$
& \textbf{FPR95}$\downarrow$
& \textbf{AUROC}$\uparrow$ \\
\midrule
RecOE (fixed rank)
& 4.51
& 98.65
& 38.46
& 90.72 \\
RecOE
& \textbf{4.36}
& \textbf{98.68}
& \textbf{36.32}
& \textbf{91.06} \\
\bottomrule
\end{tabularx}
\caption{Ablation study of progressive-rank optimization on CIFAR-10 and ImageNet-1K.}
\label{tab:ablation_progressive_rank}
\end{table}

\subsection{Experiments with an Alternative Auxiliary Dataset}
To further evaluate the generalizability of our method, we construct a more challenging setting in which both the \ind data and the auxiliary outliers are drawn from ImageNet-1K. Specifically, 100 ImageNet-1K classes are used as \ind data, while 800 disjoint classes serve as surrogate \ood data. 
We adopt the same test \ood datasets as in the setting where ImageNet-1K serves as the \ind dataset and report the averaged results across multiple datasets.
As shown in Table~\ref{tab:imagenet100}, although RecOE results in a decrease in AUROC from 85.78\% to 80.78\%, it reduces FPR95 from 69.17\% to 63.57\% and improves \ind accuracy from 77.26\% to 78.90\%. 
These results indicate that the learned subspaces retain their discriminative capability in the challenging same-source setting, despite the trade-off in AUROC, where the \ind and surrogate \ood samples are drawn from the same large-scale dataset. They further suggest that the effectiveness of RecOE does not depend solely on access to a diverse auxiliary dataset such as ImageNet-21K\_p~\cite{ridnik2021imagenet}.

\begin{table}[!ht]
\centering

{
\small

\setlength{\tabcolsep}{8pt}
\begin{tabular}{lccc}
\toprule
Methods & FPR95$\downarrow$ & AUROC$\uparrow$ & ID-ACC$\uparrow$ \\
\midrule
OE
& 69.17
& $\mathbf{85.78}$
& 77.26 \\

\ours
& $\mathbf{63.57}$
& 80.78
& $\mathbf{78.90}$ \\
\bottomrule
\end{tabular}
}
\caption{
\ood detection performance using 100 ImageNet-1K classes as \ind data and 800 ImageNet-1K classes as surrogate \ood data. 
}
\label{tab:imagenet100}
\end{table}

\subsection{Efficient variants of RecOE without Backbone Finetuning}
In OE-based methods, finetuning is applied to the entire network ($f(\cdot)$) for updating the parameters that facilitate \ood detection. 
In the optimization with  \ours given in Eq.~\eqref{eq:overall-opt} in the main paper, we introduce two matrices ${\bm U}_{\rm{in}}$ and ${\bm U}_{\rm {aux}}$  in addition to the pretrained $f(\cdot)$.
Because these learned subspaces play a central role in separating \ind and \ood features, we further investigate lightweight settings in which the backbone $g(\cdot)$ is frozen and only the subspace matrices, either alone or together with the classification head $h$, are optimized. For simplicity, we collectively denote the two subspace matrices by $\bm U$. To further simplify the optimization, these variants employ the fixed-rank subspace learning strategy. Specifically, we consider the following two lightweight variants of \ours:
\begin{itemize}
    \item \ours ($\bm U$): The subspace loss $\mathcal{L}_{\text{sub}}$ is used to only update  ${\bm U}_{\rm{in}}$ and ${\bm U}_{\rm{aux}}$, with the entire pretrained $f(\cdot)$ frozen.
    \item \ours ($h$, $\bm U$): The linear head $h$ is optimized using $\mathcal{L}_{\text{CE}}$, whereas the two basis matrices ${\bm U}_{\rm{in}}$ and ${\bm U}_{\rm{aux}}$ are optimized solely through $\mathcal{L}_{\text{sub}}$. The backbone $g(\cdot)$ remains frozen throughout training.
\end{itemize}

Table~\ref{tab:our-opt} reports the results obtained with ResNet50 on ImageNet-1K.
Both lightweight variants substantially reduce the number of trainable parameters and training time while maintaining competitive detection performance. 
Specially, \ours ($\bm U$) uses 92.4\% fewer trainable parameters and requires 89.7\% less training time per epoch than the full \ours.
Despite the slight performance drop, the lightweight variants of our proposed \ours still achieves advantageous results; for instance, simply updating $\bm U $ leads to a lower FPR95 (46.78\%) than the competitive post-hoc baseline Mahalanobis++ (49.54\%) and OE-based baseline VPC (56.50\%), and also attains higher AUROC than both.
Nevertheless, freezing the backbone $g(\cdot)$ results in a drop in {\ind} classification accuracy.
Despite this trade-off, such lightweight variants further demonstrate the flexibility and practicality of \ours, enabling its deployment in resource-constrained or time-sensitive scenarios.

\begin{table}[!htbp]
\centering
\resulttable
\begin{tabularx}{\linewidth}{>{\centering\arraybackslash}Xccccc}
\toprule
\makecell{\textbf{Updated}\\\textbf{parameters}}
& \makecell{\textbf{Time}\\\textbf{(s/epoch)}}
& \makecell{\textbf{Trainable}\\\textbf{parameters}}
& \textbf{FPR95}$\downarrow$
& \textbf{AUROC}$\uparrow$
& \textbf{ID-ACC}$\uparrow$ \\
\midrule
\ours ($\bm U$)      & 449 & 2.10M & 46.78 & 88.18 & 68.72 \\
\ours ($h$, $\bm U$) & 453 & 4.15M & 44.97 & 88.55 & 71.09 \\
\midrule
\resultrow
\ours & 4351 & 27.66M & 36.32 & 91.06 & 75.69 \\
\bottomrule
\end{tabularx}
\caption{Comparisons among variants of \ours on ImageNet-1K.
The trainable parameters and training time (s) per epoch are recorded,
together with their detection results and ID-ACC.}
\label{tab:our-opt}
\end{table}

\subsection{Hyperparameter Sensitivity}
\label{app:sensitivity}
\begin{figure*}[t]
    \centering
    \includegraphics[width=0.99\textwidth]{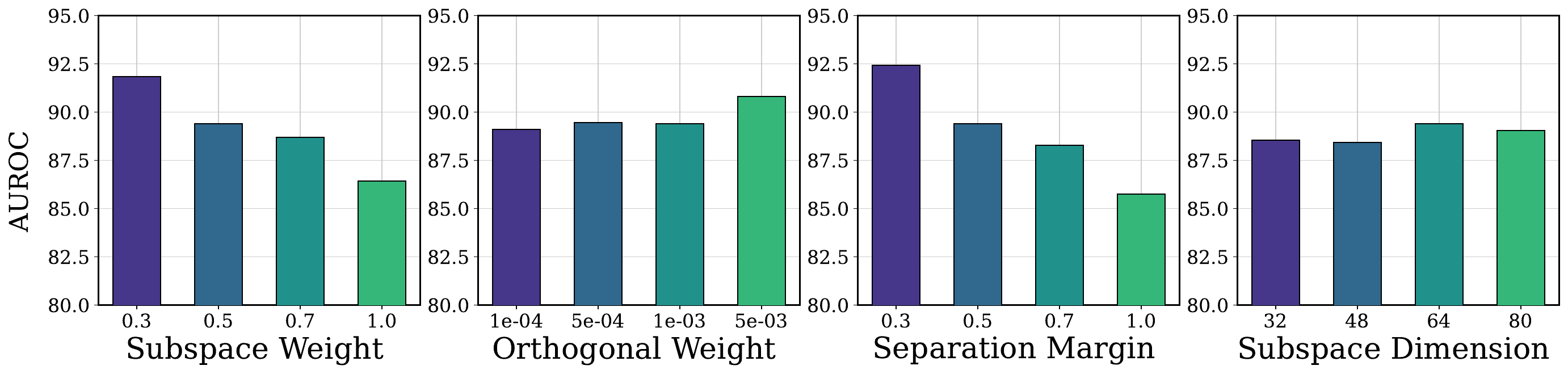} 
    \caption{Hyperparameter sensitivity analysis. From left to right: subspace-loss weight $\gamma$, orthogonality regularization weight $\beta$, separation margin $m$, and subspace dimension $k$.}
    \label{fig:ablation_hyperparams}
\end{figure*}

Using CIFAR-10 as \ind, we examine the sensitivity of \ours to the subspace-loss weight $\gamma$, orthogonality regularization weight $\beta$, separation margin $m$, and subspace dimension $k$.
We use the validation set from~\cite{hendrycks2019oe} and vary one hyperparameter at a time, keeping the others at their default values.
Figure~\ref{fig:ablation_hyperparams} shows the resulting AUROC.

\begin{itemize}
    \item \textit{Subspace weight $\gamma$:} This coefficient governs the trade-off between the classification task on \ind data  ($\mathcal{L}_{\text{CE}}$) and the subspace learning for the \ood detection task ($\mathcal{L}_{\text{sub}}$).
    Experiments suggest that an excessively large $\gamma$ may overemphasize the subspace-based objective and interfere with learning semantic representations from \ind data.

    \item \textit{Orthogonal regularization weight $\beta$:} The hyperparameter $\beta$ is related to penalizing correlation between the two subspaces, ensuring that the correspondingly learned two sets of bases capture complementary variations rather than redundant information. By reducing subspace overlap, it balances structural separation with the preservation of semantically meaningful representations.
    
    \item \textit{Subspace separation margin $m$:} The margin $m$ relates to the separation tolerance by penalizing samples whose reconstruction error relative to the opponent subspace falls below the threshold. A larger $m$ encourages a more distinct gap by pushing features apart across domains, whereas an excessively large margin can impose an overly restrictive constraint.

    \item \textit{Subspace dimension $k$:} The dimension $k$, i.e., the rank of the projection space,  determines the granularity of the captured structural information. Empirical evidence suggests that a relatively low dimensionality is more effective; it successfully distills the essential characteristics of InD data while effectively filtering out \ood noise that lacks such low-rank consistency.
\end{itemize}

\section{Conclusion and Discussion}

In this work, we move beyond the mainstream logit-output space and introduce a feature subspace learning perspective into the OE-based framework.
{Unlike previous feature-based OE methods building up the model for features with additionally prescribed structures or modules}, \ours learns two distribution-specific subspaces, ${\bm U}_{\rm{in}}$ and ${\bm U}_{\rm{aux}}$, to capture the dominant variations of \ind and \ood features while explicitly separating their distributional structures.
In inference, the contrast between the reconstruction errors induced by the two subspaces provides an effective criterion for \ood detection.
Extensive experiments across multiple benchmarks and settings further validate the effectiveness and generalizability of our method.
Overall, our method offers new insights into feature subspace learning for OE-based \ood detection, achieving substantial performance gains with only minor computational overhead.

Several aspects of the current framework warrant further investigation. RecOE uses a shared InD subspace across classes, while its detection performance has not been examined at the class level. Future work could assess class-level differences and explore whether class-aware reconstruction scores improve detection consistency. In addition, the fitting and separation objectives provide a geometric basis for reconstruction contrast, but generalization guarantees for unseen \ood distributions remain to be established. Further theoretical analysis could clarify how reconstruction margins, subspace overlap, and auxiliary-outlier coverage influence detection error.

\section*{Acknowledgments}
This work is supported by the National Natural Science Foundation of China (Grant No 62503051) and the Beijing Natural Science Foundation (L2607017).

\section*{Declaration of generative AI and AI-assisted technologies in the manuscript preparation process}

During the preparation of this work, the authors used DeepSeek for language refinement and assistance with research code implementation. After using this tool, the authors reviewed and edited the content as needed and take full responsibility for the content of the published article.

\clearpage
\appendix
\section*{Supplementary Material}

\setcounter{figure}{0}
\setcounter{table}{0}
\setcounter{equation}{0}

This supplementary material provides additional implementation details, extended experimental results and further discussions complementary to the main manuscript.

\section{Implementation Details}
\label{app:implementation}
\subsection{Training and Hyperparameter Settings}
\label{app:hyperset}
Following the setting commonly adopted in OE-based methods~\cite{hendrycks2019oe,zhu2023diversified}, we finetune pretrained backbones rather than training the complete networks from scratch.
For the CIFAR experiments, we employ a WideResNet-40-2~\cite{zagoruyko2016wide} backbone initialized from a checkpoint obtained by training on CIFAR-10 for 100 epochs on the standard classification task.
For the ImageNet-1K experiments, we utilize a ResNet-50~\cite{he2016deep} architecture initialized with standard PyTorch pretrained weights. 
During finetuning, the backbone parameters are optimized using stochastic gradient descent (SGD) with a momentum of $0.9$ and a weight decay of $5\times10^{-4}$.
The learning rate of SGD is updated at every training iteration using cosine annealing~\cite{loshchilov2017sgdr} and the initial learning rates are set to $5\times10^{-3}$ for CIFAR-10 and $1\times10^{-4}$ for ImageNet-1K.
The two orthonormal subspace bases are optimized using Riemannian Adam~\cite{becigneul2019riemannian,kochurov2020geoopt}.
Optimization details for $\bm U_{\rm {in}}$ and $\bm U_{\rm {aux}}$ are provided in ~\ref{app:progressive_subspace}.

We use mini-batch sizes of $128$ and $64$ and finetune the networks for $10$ and $4$ epochs on CIFAR-10 and ImageNet-1K, respectively.
Unless otherwise specified, the target rank of each subspace is set to $64$ for CIFAR-10 and $512$ for ImageNet-1K. For CIFAR-10, we set the separation margin to $m=0.5$, the orthogonality regularization weight to $\beta=10^{-3}$, and the subspace-loss weight to $\gamma=0.5$. For ImageNet-1K, the corresponding values are set to $m=5$, $\beta=10^{-3}$, and $\gamma=0.5$. 
Following prior work \cite{hendrycks2019oe,zhu2023diversified}, we select hyperparameters based on their overall performance across all three evaluation metrics on the validation set.

\subsection{Progressive-Rank Subspace Optimization}
\label{app:progressive_subspace}

The InD and auxiliary subspaces follow the same active-rank schedule, and their bases are initialized and expanded independently.
We describe the procedure using a generic basis $\bm U\in\mathbb R^{d\times r}$ satisfying $\bm U^\top\bm U=\bm I_r$, where $d$ is the feature dimension and $r$ is the current active rank.

\paragraph{Initialization}
Given a target rank $k\leq d$, the basis starts at rank $r^{(0)}=\max\{1,\lceil k/2\rceil\}$.
We sample a Gaussian matrix $\bm G\in\mathbb R^{d\times r^{(0)}}$ with independent standard normal entries and initialize
\begin{equation}
\bm U^{(0)}=\operatorname{qf}(\bm G),
\end{equation}
where $\operatorname{qf}(\cdot)$ returns the orthonormal factor of a reduced QR decomposition.

\paragraph{Subspace expansion}
The active rank increases every $K$ optimization steps according to Eq.~\eqref{eq:progressive_half_remaining}. For example, a target rank of $k=64$ gives the sequence
\[
32 \rightarrow 48 \rightarrow 56 \rightarrow 60
\rightarrow 62 \rightarrow 63 \rightarrow 64.
\]
When the rank increases from $r$ to $r^{+}=r+\Delta r$, where $\Delta r$ is the rank increment, we retain the current basis and sample $\bm G\in\mathbb R^{d\times\Delta r}$ with independent standard normal entries.
We project these new directions onto the orthogonal complement of the current subspace:
\begin{equation}
\widetilde{\bm G}
=
(\bm I_d-\bm U\bm U^\top)\bm G.
\end{equation}
We then orthonormalize the projected directions and construct the expanded basis:
\begin{equation}
\bm Q_{\mathrm{new}}
=
\operatorname{qf}(\widetilde{\bm G}),
\qquad
\bm U^{+}
=
\operatorname{qf}
\left(
[\bm U,\bm Q_{\mathrm{new}}]
\right).
\end{equation}
The expanded basis satisfies $(\bm U^{+})^\top\bm U^{+}=\bm I_{r^{+}}$ and contains the previously learned subspace.
All directions remain trainable after expansion, and the columns are not ordered by explained variance.

\paragraph{Optimization}
Each rank expansion changes the dimensions of the subspace parameters. We replace each Stiefel parameter with the expanded basis and reinitialize Riemannian Adam for both bases.
Optimization proceeds in stages, each with a fixed active rank $r$.
The subspace learning rate follows a cosine schedule within each stage, with maximum value
\begin{equation}
\eta_{\max}
=
\eta_U
\left(\frac{r^{(0)}}{r}\right)^{\alpha},
\label{eq:progressive_stage_lr_max}
\end{equation}
where $\eta_U$ is the initial maximum learning rate and $\alpha$ controls its decay as the rank increases. At step $\tau$ of the current stage, the learning rate is
\begin{equation}
\eta(\tau)
=
\eta_{\min}
+
\frac{1}{2}
\left(\eta_{\max}-\eta_{\min}\right)
\left[
1+\cos\left(\frac{\pi\min\{\tau,K\}}{K}\right)
\right],
\qquad \tau\geq 0,
\end{equation}
where $\tau$ counts the optimization steps since the stage began and $\eta_{\min}$ is the minimum learning rate.
The counter $\tau$ starts at zero. While $r<k$, each stage performs $K$ optimization steps with $\tau=0,\ldots,K-1$ before the next rank expansion. Each expansion starts a new stage, resets $\tau$ to zero, and updates $\eta_{\max}$ using the new active rank. Once $r=k$, no further expansion occurs, and the counter continues to increase. The learning rate follows the same schedule and remains at $\eta_{\min}$ for $\tau\geq K$.
Progressive-rank subspace optimization is summarized in  Algorithm~\ref{alg:progressive_rank}.

\begin{algorithm}[!htb]
\caption{Progressive-Rank Subspace Optimization}
\label{alg:progressive_rank}
\small\def\baselinestretch{1.05}\selectfont
\algrenewcommand{\algorithmicindent}{1.2em}
\begin{algorithmic}[1]
\Require Subspace loss $\mathcal L_{\rm sub}$; feature dimension $d$;
         target rank $k\leq d$; expansion interval $K$;
         total optimization steps $T$;
         learning-rate parameters $\eta_U,\eta_{\min},\alpha$.
\Ensure Optimized bases $\bm U_{\rm in},\bm U_{\rm aux}$.
\State $r^{(0)}\gets\max\{1,\lceil k/2\rceil\}$;
       $r\gets r^{(0)}$; $\tau\gets0$
\For{$j\in\{\mathrm{in},\mathrm{aux}\}$}
    \State Sample $\bm G_j\in\mathbb R^{d\times r}$
           with i.i.d.\ $\mathcal N(0,1)$ entries
    \State $\bm U_j\gets\operatorname{qf}(\bm G_j)$
\EndFor
\State Initialize Riemannian Adam for both bases
\For{$t=0,\ldots,T-1$}
    \If{$\tau=K$ \textbf{and} $r<k$}
        \State $\Delta r\gets
               \min\{k-r,\max\{1,\lceil(k-r)/2\rceil\}\}$
        \For{$j\in\{\mathrm{in},\mathrm{aux}\}$}
            \State Sample $\bm G_j\in\mathbb R^{d\times\Delta r}$
                   with i.i.d.\ $\mathcal N(0,1)$ entries
            \State $\widetilde{\bm G}_j\gets
                   \bm G_j-\bm U_j(\bm U_j^\top\bm G_j)$
            \State $\bm U_j\gets\operatorname{qf}
                   ([\bm U_j,\operatorname{qf}(\widetilde{\bm G}_j)])$
        \EndFor
        \State $r\gets r+\Delta r$; $\tau\gets0$
        \State Replace the Stiefel parameters
               and reinitialize Riemannian Adam
    \EndIf
    \State $\eta_{\max}\gets\eta_U(r^{(0)}/r)^\alpha$
    \State $\eta(\tau)\gets\eta_{\min}
           +\tfrac12(\eta_{\max}-\eta_{\min})
           \bigl[1+\cos(\pi\min\{\tau,K\}/K)\bigr]$
    \State Compute gradients of $\mathcal L_{\rm sub}$ with respect to both bases
    \State Update both bases by Riemannian Adam
           with learning rate $\eta(\tau)$
    \State $\tau\gets\tau+1$
\EndFor
\end{algorithmic}
\end{algorithm}

\begin{table}[!htb]
\centering
\small

\begin{tabularx}{\linewidth}{
    >{\raggedright\arraybackslash}X
    rr
}
\toprule
\textbf{Setting}
& \textbf{CIFAR-10}
& \textbf{ImageNet-1K} \\
\midrule
Target rank $k$ & 64 & 512 \\
Initial rank $r^{(0)}$ & 32 & 256 \\
Expansion interval $K$ & 500 & 1000 \\
Initial maximum learning rate $\eta_U$ & 5 & 10 \\
Minimum learning rate $\eta_{\min}$ & $10^{-3}$ & $10^{-4}$ \\
Rank-dependent decay exponent $\alpha$ & 0.5 & 0.5 \\
\bottomrule
\end{tabularx}
\caption{Progressive-rank optimization settings.
The expansion interval is measured in optimization steps,
and the settings apply to each subspace basis.}
\label{tab:progressive_settings}
\end{table}

\paragraph{Parameter settings}
Table~\ref{tab:progressive_settings} lists the subspace optimization settings.
The backbone follows its original optimization schedule.

\subsection{Computing Environment}
\label{app:env}
All experiments are conducted on a workstation running Ubuntu 24.04, equipped with an Intel Core i9-14900K CPU and two NVIDIA GeForce RTX 4090 GPUs. The CIFAR-10 experiments use a single GPU, whereas the ImageNet-1K experiments use both GPUs. The installed NVIDIA driver is version 580.142. The experiments are implemented using Python 3.7 and PyTorch 1.12 built against CUDA 11.6. Riemannian optimization of the subspace parameters is implemented using Geoopt~\cite{kochurov2020geoopt}.

\setcounter{figure}{0}
\setcounter{table}{0}
\setcounter{equation}{0}

\section{Detailed Results on Individual Datasets}
\label{app:detailedresults}

\begin{table}[htbp]
\centering
{
\scriptsize\def\baselinestretch{1.0}\selectfont
\renewcommand{\arraystretch}{1.1}
\setlength{\tabcolsep}{1.5pt}
\begin{adjustbox}{max width=\linewidth}
\begin{tabular}{@{}l*{5}{cc}@{}}
\toprule[1pt]
\multirow{3}{*}{\textbf{Method}}
& \multicolumn{10}{c}{\textbf{OoD dataset}} \\
\cmidrule(lr){2-11}
& \multicolumn{2}{c}{\textbf{Textures}}
& \multicolumn{2}{c}{\textbf{Places365}}
& \multicolumn{2}{c}{\textbf{iSUN}}
& \multicolumn{2}{c}{\textbf{LSUN\_Resize}}
& \multicolumn{2}{c}{\textbf{LSUN\_Crop}} \\
\cmidrule(lr){2-11}
& FPR95$\downarrow$ & AUROC$\uparrow$
& FPR95$\downarrow$ & AUROC$\uparrow$
& FPR95$\downarrow$ & AUROC$\uparrow$
& FPR95$\downarrow$ & AUROC$\uparrow$
& FPR95$\downarrow$ & AUROC$\uparrow$ \\
\midrule

MSP
& 59.27 & 88.50
& 60.77 & 88.13
& 56.79 & 89.53
& 53.60 & 91.02
& 30.55 & 95.68 \\

ODIN
& 54.84 & 80.28
& 48.89 & 85.66
& 28.68 & 93.85
& 23.17 & 95.05
& 9.96 & 97.90 \\

Mahalanobis
& 16.99 & 96.69
& 69.96 & 82.54
& 31.62 & 94.44
& 32.41 & 94.70
& 30.50 & 95.06 \\

Energy
& 53.17 & 85.40
& 40.22 & 89.91
& 34.60 & 92.44
& 28.48 & 93.85
& 8.12 & 98.35 \\

PCA
& 57.62 & 83.93
& 91.54 & 61.23
& 90.21 & 66.36
& 89.89 & 66.33
& 19.95 & 96.27 \\

KPCA (CoP)
& 38.30 & 91.86
& 55.98 & 85.83
& 42.83 & 91.23
& 39.10 & 92.37
& 24.89 & 95.14 \\

KPCA (CoRP)
& 36.60 & 93.49
& 41.34 & 91.05
& 31.52 & 93.64
& 25.64 & 95.02
& 23.98 & 95.80 \\

Mahalanobis++
& 15.73 & 96.99
& 45.33 & 89.19
& 18.63 & 96.67
& 16.61 & 96.99
& 1.52 & 99.69 \\

OE
& 21.81 & 96.15
& 44.86 & 91.77
& 0.00 & 100.00
& 0.00 & 100.00
& 0.99 & 99.85 \\

EnergyOE
& 33.59 & 85.47
& 58.39 & 68.32
& 0.00 & 100.00
& 0.00 & 100.00
& 0.31 & 99.83 \\

POEM
& 34.73 & 85.28
& 57.82 & 67.93
& 0.00 & 100.00
& 0.00 & 100.00
& 0.43 & 99.74 \\

MixOE
& 22.23 & 95.74
& 39.45 & 92.86
& 0.26 & 99.92
& 0.27 & 99.92
& 4.40 & 99.21 \\

DivOE
& 12.28 & 97.73
& 44.97 & 91.51
& 0.07 & 99.98
& 0.02 & 99.99
& 3.03 & 99.47 \\

GReg
& 22.28 & 93.96
& 41.18 & 91.21
& 0.00 & 100.00
& 0.00 & 100.00
& 0.00 & 100.00 \\

PFS
& 14.93 & 91.74
& 21.20 & 89.16
& 3.72 & 98.44
& 4.50 & 98.11
& 0.38 & 99.68 \\

VPC
& 38.30 & 91.22
& 29.17 & 94.16
& 4.10 & 98.50
& 3.53 & 99.00
& 0.52 & 99.70 \\

\textbf{RecOE (ours)}
& 11.32 & 96.67
& 8.27 & 97.73
& 0.78 & 99.68
& 0.62 & 99.72
& 0.82 & 99.60 \\

\bottomrule[1pt]
\end{tabular}
\end{adjustbox}
}
\caption{OoD detection performance on five OoD test datasets with CIFAR-10 as the InD data.}
\label{tab:cifar10_ood_results}
\end{table}

\begin{table}[htbp]
\centering
{
\scriptsize\def\baselinestretch{1.0}\selectfont
\renewcommand{\arraystretch}{1.1}
\setlength{\tabcolsep}{1.5pt}

\begin{adjustbox}{max width=\linewidth}
\begin{tabular}{@{}l*{4}{cc}@{}}
\toprule[1pt]
\multirow{3}{*}{\textbf{Method}}
& \multicolumn{8}{c}{\textbf{OoD dataset}} \\
\cmidrule(lr){2-9}
& \multicolumn{2}{c}{\textbf{iNaturalist}}
& \multicolumn{2}{c}{\textbf{SUN}}
& \multicolumn{2}{c}{\textbf{Places}}
& \multicolumn{2}{c}{\textbf{Textures}} \\
\cmidrule(lr){2-9}
& FPR95$\downarrow$ & AUROC$\uparrow$
& FPR95$\downarrow$ & AUROC$\uparrow$
& FPR95$\downarrow$ & AUROC$\uparrow$
& FPR95$\downarrow$ & AUROC$\uparrow$ \\
\midrule

PCA
& 95.44 & 51.98
& 97.96 & 44.83
& 97.98 & 45.15
& 46.05 & 87.79 \\

KPCA (CoP)
& 67.29 & 83.42
& 75.52 & 80.03
& 82.49 & 73.67
& 8.31 & 98.32 \\

KPCA (CoRP)
& 50.23 & 89.33
& 62.45 & 83.77
& 72.61 & 78.86
& 9.10 & 98.17 \\

Mahalanobis++
& 50.15 & 90.74
& 66.23 & 84.13
& 76.08 & 78.89
& 5.71 & 98.82 \\

GReg
& 50.42 & 91.61
& 71.38 & 81.45
& 73.97 & 79.59
& 60.67 & 81.01 \\

PFS
& 44.25 & 89.76
& 64.07 & 82.11
& 72.24 & 79.73
& 60.21 & 81.49 \\

VPC
& 43.50 & 91.20
& 60.20 & 83.10
& 66.30 & 81.50
& 56.00 & 83.00 \\

\textbf{RecOE (ours)}
& 7.85 & 98.39
& 30.67 & 91.29
& 51.69 & 85.74
& 55.08 & 88.80 \\

\bottomrule[1pt]
\end{tabular}
\end{adjustbox}
}
\caption{OoD detection performance on four OoD test datasets with ImageNet-1K as the InD data.
VPC results are taken from the original publication~\cite{xia2026vpc}; the remaining results are obtained from our experiments.}
\label{tab:imagenet_ood_results}
\end{table}

In the main paper, we report the averaged OoD detection performance and standard deviations across multiple datasets for concise comparison. 
To provide a more detailed analysis of the effectiveness of RecOE under diverse OoD scenarios, we further report results for each OoD test dataset in this supplementary material. The detailed results on the CIFAR-10 and ImageNet-1K benchmarks are presented in Tables \ref{tab:cifar10_ood_results} and \ref{tab:imagenet_ood_results}, respectively.

Following the experimental settings of DivOE~\cite{zhu2023diversified}, we reproduce PCA~\cite{guan2023revisit}, KPCA (CoP and CoRP)~\cite{fang2024kernel}, Mahalanobis++~\cite{mueller2025mahalanobispp}, GReg~\cite{sharifi2024gradient}, and PFS~\cite{wu2025pfs}.
For CIFAR-10, we additionally reproduce VPC~\cite{xia2026vpc}; results for the remaining baselines are taken from DivOE.
For ImageNet-1K, Table~\ref{tab:imagenet_ood_results} reports our reproduced baseline results and the published VPC results.

These detailed results highlight the substantial performance gains of our proposed framework over representative baselines across varied OoD datasets. Notably, RecOE significantly reduces the FPR95 on complex OoD datasets where the compared methods typically struggle, such as Textures\cite{cimpoi2014describing} and Places365\cite{zhou2016places} for CIFAR-10, and iNaturalist\cite{van2018inaturalist} for ImageNet-1K. 
While OE-based frameworks often exhibit slight performance trade-offs across different datasets, our approach achieves a better overall balance.
For example, although RecOE performs slightly worse than some OE baselines on iSUN and LSUN\_Resize, where performance is already near saturation, it achieves better overall performance across the full set of OoD datasets.







\bibliographystyle{elsarticle-num}
\bibliography{reference}

\end{document}